\documentclass[11pt]{article}

\usepackage[T1]{fontenc}
\usepackage[utf8]{inputenc}
\usepackage[margin=1in]{geometry}
\usepackage{natbib}
\usepackage{amsmath,amssymb,booktabs,array,calc,graphicx}
\usepackage{xurl,hyperref}
\usepackage{microtype}
\usepackage{textcomp}
\hypersetup{colorlinks=true,allcolors=blue}
\title{Neuroevolution Arena: Nested Ecological Evaluation of Update-and-Inheritance Regimes across Neural Architectures}

\author{%
  Yuxu Ge\\
  University of York, United Kingdom\\
  \href{mailto:research@yuxu.ge}{research@yuxu.ge}\\
  \href{https://orcid.org/0009-0008-2990-4886}{ORCID: 0009-0008-2990-4886}
  \and
  Yifei Cheng\\
  University of York, United Kingdom\\
  \href{mailto:mlq535@york.ac.uk}{mlq535@york.ac.uk}\\
  \href{https://orcid.org/0009-0005-3914-5172}{ORCID: 0009-0005-3914-5172}%
}
\date{}

\begin{document}
\maketitle

\begin{abstract}
Competitive artificial-life systems can rank trained controllers differently under training and ecological evaluation. We present Neuroevolution Arena, a GPU-accelerated spatial ecology of independently parameterized neural-network cells, and an audit-tracked nested evaluation protocol. Three implementation-specific update-and-inheritance regimes---EvoEvo, EvoRL, and RLRL---are crossed with two neural architectures for 50,000 generations in three independent training runs per condition. One saved elite-controller artifact from each of the 18 runs enters an aligned-run frozen-evaluation design comprising 198 computational jobs. Pairwise effects average three seed-defined ecological contexts---two cooperation-permitting and one attack-permitting---within each aligned training-run block; the independent level remains n = 3 runs per condition. RL-enabled regimes attain higher recorded training fitness than EvoEvo, whereas pairwise outcomes show architecture-conditioned majority patterns and substantial artifact dependence. Six-way winners vary across artifacts and contexts, and the prespecified survival endpoint has a complete floor. We contribute a nested protocol that separates training-run artifacts from evaluation contexts and exposes, rather than conceals, their different sources of variation.
\end{abstract}

\begin{center}
\small Preprint. This manuscript has not undergone journal peer review.
\end{center}


\medskip
\noindent\textbf{Data and trained models:} \url{https://huggingface.co/datasets/geyuxu/alife2026-data} and \url{https://huggingface.co/geyuxu/alife2026-model}

\section{Introduction}\label{introduction}

Biological evolution changes inherited structure across generations, whereas lifetime learning changes behavior within an organism. \citet{baldwin1896} proposed that plasticity can alter selection, and \citet{hinton1987} modeled one computational route by which learning can guide evolution. This relationship motivates studying inherited and lifetime-updated neural components together, without assuming that every prescribed combination implements a Baldwin Effect. In artificial systems, neural architecture search commonly optimizes topology on fixed benchmarks, reinforcement learning (RL) optimizes policies within a chosen architecture, and neuroevolution searches inherited controllers. Their effects are often reported separately. We instead ask whether performance rankings in a shared artificial ecology vary jointly with network architecture and an implementation-specific update-and-inheritance regime (hereafter, regime).

This issue becomes especially important in competitive ecological settings, where evaluation is endogenous even when cell-network weights are held fixed. Success is reflected in outcomes such as pairwise population persistence, multi-way coexistence or exclusion, and recovery after population collapse. During frozen evaluation, policies do not learn or evolve, but population composition, spatial contact, reproduction, and resource pressure continue to change as consequences of those fixed weights. Benchmark superiority therefore need not imply ecological robustness. Match structure, saved training-run artifact, and seed-defined ecological context can all influence which strategy appears strongest. This motivates an evaluation design that covers every available training run, separates pairwise from multi-way questions, and preserves the nesting of contextual repetitions within trained models.

To investigate this question, we present Neuroevolution Arena, a GPU-accelerated artificial life platform in which up to 10,000 cells, each equipped with an independent neural network, compete on a toroidal grid. Unlike standard Neural Cellular Automata (NCA) \citep{mordvintsev2020}, where all cells share a single learned update rule, our system supports heterogeneous populations: cells may differ in network architecture and in the update-and-inheritance rules applied to W1 and W2. EvoEvo applies evolutionary crossover and mutation to both layer groups, EvoRL combines W1 evolution with lifetime RL on W2, and RLRL applies lifetime RL to both while retaining reproductive inheritance and selection. We train a replicated 3 $\times$ 2 regime--architecture factorial and evaluate one saved run-level controller artifact from every training run through audit-tracked frozen competition. Pairwise effects are estimated from population area under the curve (AUC) in an \emph{aligned-run} design: artifacts with the same run index compete, and three seed-defined ecological contexts are nested within that block. Six-way and survival protocols probe complementary ecological properties without being pooled into a single score.

We ask two questions. \textbf{RQ1:} How are regime and architecture jointly associated with recorded training fitness and frozen ecological outcomes in the tested system? \textbf{RQ2:} How do those outcomes vary across saved training-run artifacts, opponent sets, and seed-defined ecological contexts? The contribution is methodological and empirical: the study distinguishes independent training replication from repeated evaluation of a saved artifact, reports a non-resolving primary endpoint as a null result, and identifies architecture-conditioned patterns without promoting any condition to a universal winner. A separate, unmatched training-horizon dataset is retained only as an exploratory appendix and does not answer either main research question.

\section{Related Work}\label{related-work}

This section sits at the intersection of four research threads: neural cellular automata, neuroevolution, the Baldwin Effect, and neural architecture evaluation. We review each in turn, focusing on the conceptual background and open questions that motivate the present study.

\subsection{Neural Cellular Automata}\label{neural-cellular-automata}

Neural Cellular Automata (NCA) replace hand-crafted CA rules \citep{langton1986} with learned neural network update functions. \citet{mordvintsev2020} demonstrated that a small convolutional network could learn local update rules that produce self-organizing, self-repairing morphogenesis from a single seed cell, an influential result showing that complex global patterns can emerge from simple learned local rules. Subsequent work extended NCA to 3D structure generation \citep{sudhakaran2021}, steerable orientation \citep{randazzo2023steerable}, and texture synthesis \citep{niklasson2021}. A unifying feature of these systems is that all cells share a single set of learned parameters, optimized end-to-end via backpropagation through time. This weight-sharing constraint promotes coherent global behavior but limits the kind of intra-population diversity that characterizes biological ecosystems.

Recent efforts have begun introducing ecological dynamics into NCA frameworks. Biomaker CA \citep{randazzo2023biomaker} supports plant-like organisms that grow, reproduce with mutation, and compete for nutrients in a shared grid, using meta-evolution to discover viable morphogenetic programs. Notably, Biomaker CA allows different organisms to have different learned programs, introducing a form of heterogeneity within the NCA paradigm; however, this heterogeneity operates primarily at the level of organism-level programs rather than per-cell independent neural parameters within a single population. Petri Dish NCA \citep{zhang2025} goes further by allowing multiple independently parameterized NCA agents to coexist and adapt via continual gradient-based learning within a shared substrate, demonstrating that ecological dynamics can emerge even when agents optimize via gradient descent. Relative to Biomaker CA, this places more emphasis on parameter granularity within the shared substrate, but it still centers ecological adaptation around gradient-based optimization rather than comparing multiple optimization paradigms within the same competitive environment. Concurrently, systems like Lenia \citep{chan2019} and Flow-Lenia \citep{plantec2025} explore continuous CA that produce rich emergent dynamics---including self-replicating patterns and evolution-like adaptation---without neural networks, relying instead on parametric kernel functions and conservation laws. These systems are closely related in ecological intuition, but they address it through a different modelling paradigm: they study emergence and adaptation in continuous cellular systems rather than heterogeneous neural agents with learned update rules.

This lineage leaves several related questions open. First, while recent work has introduced heterogeneity into ecological CA systems, less attention has been given to settings in which each cell maintains its own neural network weights and the substrate is heterogeneous at the individual level. Second, existing differences also fall at the level of optimization mechanism: many systems rely primarily on gradient-based training, meta-evolution, or a single optimization regime, leaving less work on how evolution and reinforcement learning interact within the same competitive setting. Third, learning paradigms and network architectures are typically studied separately, whereas a shared ecological setting makes it possible to compare them directly within the same environment. More broadly, these gaps motivate examining NCA-based systems with per-cell independent neural networks and multiple learning paradigms operating simultaneously within a single simulation.

\subsection{Neuroevolution}\label{neuroevolution}

Neuroevolution optimizes neural network parameters or topologies through evolutionary algorithms. NEAT \citep{stanley2002} co-evolves weights and topology with speciation to protect structural innovations. Evolution Strategies \citep{salimans2017} showed that gradient-free optimization of fixed-topology networks can match RL on standard benchmarks, while deep genetic algorithms scaled population-based search to large neural policies \citep{such2017}. Novelty search further demonstrated the value of behavioral exploration when explicit objectives are deceptive \citep{lehman2011}. Weight Agnostic Neural Networks \citep{gaier2019} showed that topology alone can encode useful inductive biases. Together, these studies establish neuroevolution as a flexible alternative to gradient-based optimization over parameters, topologies, or both.

Neuroevolution also supports relational fitness defined through changing opponents rather than a fixed objective. \citet{sims1994} evolved virtual creatures with coupled morphology and control in competitive tasks, and \citet{rosin1997} developed methods for addressing pathologies in competitive coevolution. \citet{stanley2004} later used competitive coevolution in a robot-duel domain to study controller complexification. Open-ended-evolution research frames sustained novelty as a central evaluation challenge \citep{packard2019}, while POET pairs environmental challenge generation with agent optimization and transfer \citep{wang2019}. These systems differ from the persistent spatial ecology studied here, but they establish opponents and environments as part of the evaluation context rather than a fixed test set.

Even so, much of the neuroevolution literature evaluates agents against relatively explicit objectives, whether in fixed benchmarks, structured competitive tasks, or pairwise contests with clearly specified success criteria. Persistent ecological settings receive comparatively less attention even though success there is shaped jointly by energy acquisition, reproduction, survival, and interaction with other changing populations. The gap motivating this study is therefore not simply which optimizer scores highest on a predefined task, but how alternative optimization regimes persist and compete over time while inhabiting and jointly shaping the same shared ecology.

\subsection{The Baldwin Effect}\label{the-baldwin-effect}

\citet{baldwin1896} proposed that an organism's capacity for lifetime learning can guide the course of evolution by smoothing the fitness landscape, making otherwise inaccessible genotypes reachable through selection. \citet{hinton1987} provided the first computational demonstration: in a needle-in-a-haystack fitness landscape where the chance of randomly finding the optimal genotype was astronomically small, populations with learnable alleles converged on the optimal genotype far faster than non-learning populations. The key insight was that learning allows organisms to ``discover'' nearby fitness peaks during their lifetimes, creating a fitness gradient that evolution can follow across generations.

Subsequent work explored the Baldwin Effect under varying conditions: dynamic environments where the fitness landscape changes over time \citep{suzuki2004}, different plasticity models that constrain what can be learned \citep{morgan2020}, and evolving learning rates that allow the balance between genetic and learned behavior to itself be optimized \citep{bull1999}. A recent revival of interest has re-examined the original Hinton-Nowlan model, demonstrating that the magnitude of the Effect depends sensitively on the nature of plasticity and fitness landscape structure \citep{fontanari2017}. However, most computational studies of the Baldwin Effect use simple genetic algorithms with bit-string genotypes in well-mixed populations with static or slowly changing fitness landscapes.

These studies motivate the layer assignment examined here, but they do not imply that a prescribed split between evolutionary and lifetime updates constitutes a demonstrated Baldwin Effect. The present experiments therefore ask how three implementation-specific update-and-inheritance combinations behave under a competitive spatial ecology, and treat Baldwinian dynamics as a hypothesis requiring separate causal ablations.

In the present study, the Baldwin Effect serves only as one motivation for separating evolved and lifetime-updated layer groups. EvoRL assigns evolutionary crossover and mutation to W1 and lifetime RL to W2. In the 50K cross-experiment, W2 weight updates are disabled for the first 3\%, although reward state can accumulate; W1 evolution remains active, and W1 evolution and W2 RL continue concurrently after onset. This is not a completed evolutionary phase followed by a separate RL phase. EvoEvo applies evolution to both layer groups. RLRL applies lifetime RL to both layer groups but still includes reproduction, selection, parent-weight averaging, and output-layer reset. The comparison therefore tests three implementation-specific update-and-inheritance regimes rather than pure, isolated optimization algorithms.

\subsection{Neural Architecture Evaluation}\label{neural-architecture-evaluation}

Neural Architecture Search (NAS) automates the design of network topologies and spans reinforcement-learning, differentiable, and evolutionary search strategies \citep{elsken2019}. \citet{zoph2017} used reinforcement learning to search over architectures, DARTS introduced differentiable relaxations \citep{liu2019}, and regularized evolution demonstrated a competitive evolutionary alternative \citep{real2019}. These methods generally evaluate candidate architectures on fixed benchmarks and rank them by held-out performance.

A growing body of work questions whether benchmark performance captures the full picture. Architectures that perform well on standard tasks may fail under distribution shift \citep{recht2019}, and rankings can change substantially with training duration, hyperparameter choices, and random seed \citep{dodge2020}. These findings suggest that architecture evaluation may be more context-dependent than commonly assumed---the apparent ``best'' architecture is not simply an intrinsic property of the topology but also depends on the evaluation conditions. In particular, \citet{dodge2020} showed that different random seeds and early stopping criteria can substantially change fine-tuning outcomes for the same model, raising questions about the reproducibility of evaluation results even within the controlled setting of supervised learning.

These concerns extend naturally to competitive ecological evaluation. Rankings may be still more context-dependent when they emerge from interacting opponents and endogenous population dynamics rather than a fixed test set. In the frozen protocol used here, controller parameters remain fixed, but reproduction, spatial contact, population fragmentation, and resource pressure continue. The broader question is therefore whether architecture rankings remain stable across saved training-run artifacts, opponent sets, and ecological contexts. This motivates treating architecture as a factor whose apparent value may vary with evaluation context rather than assuming a universally best topology.

\section{System: Neuroevolution Arena}\label{system-neuroevolution-arena}

We describe the main components of Neuroevolution Arena, implemented in PyTorch. Dominant grid-state and per-cell network operations use batched tensors on the selected PyTorch device; topology and speciation routines and some control paths use Python and CPU processing. The formal frozen-evaluation runner required CUDA. No hardware-normalized throughput claim is made because this paper does not report a performance benchmark.

\subsection{Environment}\label{environment}

The world is a 100 $\times$ 100 occupancy grid with 10,000 possible cell locations. Each location is empty or contains one living cell. A living cell stores energy, hunger, age, a two-dimensional velocity, explicit lineage metadata, chemical affinity, a cooperation scalar, independently maintained action-network weights, and independently maintained context-head weights. Energy is changed by global metabolism, reproduction cost, internal-cell harvesting and sharing, pioneer subsidy, feeding, movement, and combat transfer. Most local tensor rolls and cell movement wrap toroidally; chemical diffusion and the internal-energy border convolution use zero padding.

Cells incur a density-dependent metabolic cost intended to limit unbounded population growth. The multiplier is 1 below a population of 800 and increases quadratically only for excess population. At 2,000 cells it is approximately 86.2$\times$:

\begin{equation}\label{eq:metabolism}
c = c_0\left[1 + \left(\frac{\max(0,p-800)}{130}\right)^2\right],
\end{equation}

where $p$ is the total number of living cells across all species and $c_0 = 0.05$ is the baseline metabolic rate per step. The reference point of 800 cells (approximately 8\% grid occupancy) and the 130-cell scaling parameter were selected through pilot tuning to keep small populations viable while making crowding increasingly costly. Together they approximate a soft carrying-capacity pressure: populations near or below 800 experience near-baseline metabolism, while populations exceeding this threshold face quadratically increasing energy drain.

Cell mortality is governed by two accumulating pressures. First, every living cell's hunger increases unconditionally each step by a base increment that accelerates with age:

\[\Delta h_{\mathrm{base}} = 1 + \left\lfloor \frac{\mathrm{lifetime}}{500} \right\rfloor.\]

where lifetime is the cell's age in simulation steps. This aging term gives even well-positioned cells a finite expected lifetime, reducing the possibility of effectively immortal lineages. Second, Conway-style local density rules \citep{gardner1970} impose additional hunger penalties based on conspecific (genome distance \textless{} 0.5) neighborhood counts within the Moore neighborhood. Isolated cells with fewer than 2 conspecific neighbors receive +10 hunger per step. Overcrowded cells accumulate hunger proportional to the excess beyond a threshold of 6 neighbors:

\[\Delta h_{\mathrm{crowd}} = 10\max(0,n_{\mathrm{kin}}-6).\]

Local density penalties apply after a cell's first two lifetime steps. Cells die when hunger reaches 150 or energy reaches zero. A pioneer subsidy applies when dynamic \texttt{species\_id} population is at most 3: depending on neighborhood openness, it removes up to 8 hunger and adds up to 0.15 energy plus a refund of density-dependent metabolism. This mitigates but does not exempt a mature isolated pioneer from the base and +10 isolation increments; at age below 500, a fully open isolated pioneer has a net hunger increment of approximately +3 per step.

Three additional birth and energy mechanisms remain active in training and frozen evaluation. First, an empty location whose stored genome is within the 0.5 threshold of exactly three living neighbors can receive a Conway-style clone, subject to a three-step frontier-death cooldown. Second, every 10 steps an enclosed empty location with at least six similar living neighbors can be filled from an explicit donor; frozen evaluation uses an exact clone rather than neighbor averaging. Third, a cell surrounded by eight chemical-affinity-similar neighbors gains 0.5 energy and reduces hunger by 5, while adjacent border cells gain 0.1 energy per qualifying internal neighbor. These operational kin rules can cross model-origin boundaries when affinities are standardized.

\subsection{Chemical Signaling}\label{chemical-signaling}

A four-channel chemical field overlays the grid and contributes to neural input, but not directly to the combat rule. Each channel is convolved with a fixed 7 $\times$ 7 Gaussian-like kernel using zero padding. The original and convolved fields are blended with diffusion rate 0.3, followed by 2\% global decay per step. Living cells with positive inherited chemical affinity secrete at rate 0.1 $\times$ affinity $\times$ clamp(energy/100, 0, 1), and every channel is clamped to {[}0, 10{]}. These constants are fixed across conditions; the study does not separately identify a functional role for the resulting fields.

For chemical direction sensing, the four channels are first summed. In each of eight directions, the cell samples distances 1--5 with weights 1, 0.5, 0.25, 0.125, and 0.0625, then multiplies the result by 0.2. The four local channel concentrations are also multiplied by 0.2 and supplied separately. These twelve values are sensory inputs only. Contact outcomes instead use the cooperation scalar, energy, explicit species identity, and underdog protection described in Section 3.5.

\subsection{Per-Cell Neural Architecture}\label{per-cell-neural-architecture}

Each cell contains its own action network, with independently maintained W1 and W2 parameters rather than population-wide weight sharing. The per-cell network consists of a perception layer (W1) and a three-layer decision network (W2), takes a 26-dimensional input vector, and produces probabilities for 11 discrete actions. A separate per-cell context head receives six statistics aggregated by explicit species identity and supplies a four-dimensional strategy input; cells in one species share the aggregate statistics, not one guaranteed-identical set of context-head weights.

Perception layer (W1): Maps raw input to a hidden representation via a single linear layer with tanh activation, compressing local sensory inputs into an intermediate state that serves as the cell's perceptual representation:

\begin{equation}\label{eq:perception}
\mathbf{h}=\tanh(\mathbf{x}W_1),\qquad W_1\in\mathbb{R}^{26\times d_h},
\end{equation}

where $d_h$ is the hidden dimension, which varies across architecture configurations.

Skip connection and context injection: The hidden activation h is concatenated with the raw input x and a four-dimensional cell-local strategy vector s. The vector s is computed from species-aggregate statistics by that cell's inherited context-head weights (Section 3.6), yielding:

\begin{equation}\label{eq:concat}
\mathbf{z}=[\mathbf{h};\mathbf{x};\mathbf{s}]\in\mathbb{R}^{d_h+30}.
\end{equation}

The skip connection keeps raw sensory input available after perceptual processing, while s supplies low-dimensional conditioning on the current aggregate state of the cell's explicit species. Because context-head parameters are stored per cell, s need not be identical for all cells assigned the same \texttt{species\_id}.

Decision network (W2): A 3-layer MLP that maps the combined vector to action probabilities, with the earlier layers transforming the integrated perceptual and strategy signal and the final layer producing the action distribution:

\begin{equation}\label{eq:decision}
W_{2a}:d_z\!\to\!d_r,\qquad W_{2b}:d_r\!\to\!d_r,\qquad W_{2c}:d_r\!\to\!11.
\end{equation}

with tanh activations after W2a and W2b, and softmax after W2c.

The 26-dimensional input vector contains: (i) eight neighbor energies divided by 100 (dimensions 0--7); (ii) similar- and different-neighbor counts divided by 8 (8--9), where similarity for this input uses the four chemical-affinity values and the 0.5 distance threshold; (iii) own energy divided by 100 (10); (iv) total neighbor count divided by 8 (11); (v) eight directional chemical signals multiplied by 0.2 (12--19); (vi) four local chemical concentrations multiplied by 0.2 (20--23); and (vii) velocity divided by the configured maximum movement step, 3 (24--25). Energy values are not clipped by this preprocessing. All conditions use the same encoding, but no scale-invariance claim is made.

The 11 actions are eight directional reproduction attempts, stay, self-destruct, and feed. Stay skips an action-specific reproduction or transfer; it does not itself award energy, although global ecological updates still apply. Self-destruct kills the cell and distributes its remaining energy among adjacent cells whose four-dimensional chemical-affinity distance is below 0.5. Feed transfers half the cell's energy among neighbors passing the same affinity rule while the donor remains alive. There is no action-network output labelled move: displacement is applied separately from the two-dimensional context movement head through the Boid-style update \citep{reynolds1987} in Section 3.6.

Architecture bucketing. Training conditions are assigned predefined architecture descriptors specifying $d_h$ (perception width), $d_r$ (decision width), and whether skip connections are enabled. Architecture functions as an inherited lineage-level property within a run and as an experiment-level condition across runs. The corrected frozen evaluation directly compares Baseline64 (hereafter MLP4) and Wide128; Hybrid64-128 and NoSkip64 appear only in the unmatched training-only appendix. Table~\ref{tab:1} summarizes all four descriptors while preserving this evidence boundary.

\begin{table*}[t!]
\centering
\small
\caption{Architecture configurations tested. Counts cover the per-cell W1 and W2 weight matrices; these matrices have no bias terms. The separate context strategy and movement heads are common in shape across architectures and are excluded. ``MLP4'' is retained as the legacy name for the Baseline64 controller. ``Hybrid64-128'' replaces the misleading legacy label ``Deep64'': it has the same matrix depth but uses $d_h=64$ and $d_r=128$.}\label{tab:1}
\resizebox{\textwidth}{!}{%
\begin{tabular}{@{}
  >{\raggedright\arraybackslash}p{(\linewidth - 8\tabcolsep) * \real{0.3077}}
  >{\centering\arraybackslash}p{(\linewidth - 8\tabcolsep) * \real{0.1538}}
  >{\centering\arraybackslash}p{(\linewidth - 8\tabcolsep) * \real{0.2923}}
  >{\centering\arraybackslash}p{(\linewidth - 8\tabcolsep) * \real{0.1154}}
  >{\centering\arraybackslash}p{(\linewidth - 8\tabcolsep) * \real{0.1308}}@{}}
\toprule\noalign{}
\begin{minipage}[b]{\linewidth}\raggedright
\textbf{Architecture}
\end{minipage} & \begin{minipage}[b]{\linewidth}\centering
\textbf{W1}
\end{minipage} & \begin{minipage}[b]{\linewidth}\centering
\textbf{W2 path}
\end{minipage} & \begin{minipage}[b]{\linewidth}\centering
\textbf{Skip}
\end{minipage} & \begin{minipage}[b]{\linewidth}\centering
\textbf{Action-network weights}
\end{minipage} \\
\midrule\noalign{}
Baseline64 (MLP4) & 26$\rightarrow$64 & 94$\rightarrow$64$\rightarrow$64$\rightarrow$11 & Yes & 12,480 (\textasciitilde12.5K) \\
Wide128 & 26$\rightarrow$128 & 158$\rightarrow$128$\rightarrow$128$\rightarrow$11 & Yes & 41,344 (\textasciitilde41.3K) \\
Hybrid64-128 (legacy: Deep64) & 26$\rightarrow$64 & 94$\rightarrow$128$\rightarrow$128$\rightarrow$11 & Yes & 31,488 (\textasciitilde31.5K) \\
NoSkip64 & 26$\rightarrow$64 & 68$\rightarrow$64$\rightarrow$64$\rightarrow$11 & No & 10,816 (\textasciitilde10.8K) \\
\bottomrule
\end{tabular}
}
\end{table*}

\subsection{Update-and-Inheritance Regimes}\label{update-and-inheritance-regimes}

The W1/W2 separation defines three regimes. Each cell carries a model-type attribute, orthogonal to architecture, that controls which layer groups receive lifetime RL, evolutionary crossover or mutation, parental averaging, and output-layer reset.

EvoEvo. Both W1 and W2 are inherited through evolutionary reproduction. Under sexual reproduction, the two layer groups use crossover and mutation; no lifetime RL update is applied. The W2c output layer is not randomly reset for EvoEvo. Adaptive change therefore occurs through inherited variation and population selection across generations.

RLRL. Both W1 and W2 receive a custom reward-modulated, backpropagation-like update during a cell's lifetime; the implementation is not standard REINFORCE. For the output matrix W2c, the direct term is:

\begin{equation}\label{eq:rl-update}
\Delta W=\eta r\,\mathbf{h}_{\mathrm{in}}^{\mathsf{T}}\mathbf{a}_{\mathrm{onehot}},
\end{equation}

where $\eta = 0.01$ is the learning rate, $r$ is the implementation-defined scalar reward, $\mathbf{h}_{\mathrm{in}}$ is W2c's input activation, and $\mathbf{a}_{\mathrm{onehot}}$ is the selected action. Updates to W2b, W2a, and, where enabled, W1 include tanh derivatives and downstream-weight terms. Reward increments are +1.5 for successful reproduction, +0.5 for frontier reproduction, +1.0 for a pure combat win, +0.3 for choosing a foreign-facing direction with at least three same-species neighbors, +0.05 for a direction without a foreign neighbor, and +0.15 for maintaining two to four same-species neighbors; failed reproduction and solo foreign-facing choices incur -0.3 and -0.1. Replay is maintained per architecture with capacity 10,000 and batch size 256; samples are filtered by model type and their mean update is broadcast to alive cells of that architecture and model type. Thus, ``lifetime RL'' names when experiences are collected, not an isolated buffer and update unique to each cell. RLRL retains reproduction and selection: sexual offspring average parental W1 and W2 internal layers, while W2c is randomly reinitialized. Evolutionary crossover and mutation are disabled for its action network, but its cooperation scalar is still inherited with Gaussian birth noise in the legacy training path.

EvoRL. W1 uses evolutionary crossover and mutation, while W2 receives lifetime RL. Sexual offspring directly average parental W2a/W2b values after any RL modification and randomly reinitialize W2c; this direct inheritance of acquired internal weights is Lamarckian-like rather than a classical Baldwin mechanism. In the cross-experiment, W2 weight updates begin at 3\% of the 50K horizon and W1 evolution continues thereafter. However, a known warmup defect leaves accumulated reward uncleared during this interval, so surviving cells can carry pre-onset reward into the first RL update; the transition is not a clean reset between phases. EvoRL is thus a layer-partitioned concurrent regime after a short update-free interval for W2, not a sequential evolution-then-RL schedule. Its relation to the Baldwin Effect is motivational rather than a direct operational demonstration.

Table~\ref{tab:2} summarizes the update, inheritance, and RL-onset differences.

\begin{table*}[t!]
\centering
\small
\caption{Regime summary. ``Evolution'' denotes crossover/mutation during reproduction; ``RL'' denotes within-lifetime updating. RLRL retains reproductive averaging and selection despite having no evolutionary crossover or mutation.}\label{tab:2}
\resizebox{\textwidth}{!}{%
\begin{tabular}{@{}
  >{\raggedright\arraybackslash}p{(\linewidth - 6\tabcolsep) * \real{0.1923}}
  >{\centering\arraybackslash}p{(\linewidth - 6\tabcolsep) * \real{0.2692}}
  >{\centering\arraybackslash}p{(\linewidth - 6\tabcolsep) * \real{0.2692}}
  >{\centering\arraybackslash}p{(\linewidth - 6\tabcolsep) * \real{0.2692}}@{}}
\toprule\noalign{}
\begin{minipage}[b]{\linewidth}\raggedright
\textbf{Regime}
\end{minipage} & \begin{minipage}[b]{\linewidth}\centering
\textbf{W1 update / inheritance}
\end{minipage} & \begin{minipage}[b]{\linewidth}\centering
\textbf{W2 update / inheritance}
\end{minipage} & \begin{minipage}[b]{\linewidth}\centering
\textbf{RL onset}
\end{minipage} \\
\midrule\noalign{}
EvoEvo & Evolution & Evolution; no W2c reset & never \\
RLRL & Lifetime RL + parent averaging & Lifetime RL + averaging; W2c reset & generation 0 \\
EvoRL & Evolution throughout & Lifetime RL + averaging; W2c reset & 3\% of training \\
\bottomrule
\end{tabular}
}
\end{table*}

\subsection{Reproduction, Selection, and Combat}\label{reproduction-selection-and-combat}

A cell may attempt reproduction by selecting one of eight directions when energy is at least 8. The target must be empty and pass the soil-cooldown rule. Conditional on availability, success probability is 0.7 when the target has at most two genome-similar neighbors and 0.5 otherwise. A successful parent pays a base cost of 1 energy, multiplied by max(1, dynamic-species population/40), while the offspring is initialized with 10 energy; energy is intentionally not conserved. The intended training rule searches adjacent cells for a mate with full 12-dimensional genome distance below 0.5 and falls back to asexual reproduction for a different architecture. EvoEvo uses binary crossover and Gaussian noise with scale 0.1 for W1 and all W2 matrices, including W2c. EvoRL applies crossover/noise to W1, averages parental W2a/W2b under sexual reproduction, and randomly initializes W2c at every birth. RLRL averages sexual W1/W2a/W2b, applies no evolutionary action-network noise, and also randomly initializes W2c. Asexual inheritance follows the corresponding regime rules. A known legacy defect marks the target alive before mate search, so a not-yet-initialized target can be selected using stale state; saved training artifacts therefore reflect this implemented, potentially contaminated mate-selection path. Formal frozen evaluation avoids this branch and makes every birth an exact single-parent clone of controller and lineage-template state.

The implementation uses three related identity representations. \texttt{model\_origin} is an immutable condition label used to aggregate competition AUC. A dynamic integer \texttt{species\_id} controls species aggregation, contact rules, pioneer counts, and several lineage operations. It is inherited at birth, but every 50 steps each disconnected eight-neighbor component is split into a new child \texttt{species\_id}; thus one model origin can contain many ecological species and same-origin fragments can be foreign for contact rules. Finally, each cell has a 12-dimensional genome proxy: four raw statistics for W1 and four for flattened W2 (mean, standard deviation, mean absolute value, and maximum absolute value), plus four chemical-affinity values. Full-genome Euclidean distance below 0.5 defines mating compatibility and density-based kin counts. Chemical-affinity distance alone defines the action-network kin counts, self-destruct and feed recipients, and internal-energy neighborhoods. The weight statistics are not normalized across architectures, so the threshold may have architecture-dependent scale effects. These are operational labels and similarity rules, not a validated biological species definition.

Contact rules operate between adjacent cells with different explicit \texttt{species\_id} values and do not use chemical concentration. Each cell carries a cooperation scalar: values above zero select cooperate and values at or below zero select attack. If both attack, the higher-energy cell wins; ties give each cell an independent 0.5 death draw. A cell whose species population is less than half that of the neighbor is protected from combat death. An attacker beats a cooperator, while two cooperators normally cause the lower-energy cell to adopt the stronger neighbor's species. Energy from cells killed in a step is pooled among pure winners and one winner offspring may occupy a death site. During formal frozen evaluation, species conversion by cooperation is disabled together with other heritable-state changes.

During training, dominance control samples up to 500 full 12-dimensional genomes, connects pairs whose distance is below 0.5, and treats graph-connected components as genome clusters. All cells are assigned to the nearest sampled cluster center within the same threshold. If one cluster exceeds 50\% of the total population, the rule invokes an interleaved mutation. The affected action-network layers remain descriptor-specific: EvoEvo can mutate W1 and W2, EvoRL W1 only, and RLRL neither; associated affinity, cooperation, and context-head mutation is likewise restricted to model types with at least one evolutionary action-network layer. This is a diversity heuristic defined by genome clustering rather than dynamic \texttt{species\_id}. It is disabled in formal frozen evaluation, where dominance, fission, and other controller mutations are audited as zero.

\subsection{Species-Aggregated Context Heads and Boid Movement}\label{species-aggregated-context-heads-and-boid-movement}

The simulator first aggregates six statistics by dynamic \texttt{species\_id}: population divided by grid area, mean energy divided by \texttt{MAX\_ENERGY} $= 10^{9}$, mean full-genome-similar neighbor count divided by 8, mean hunger divided by 150, frontier ratio, and crowded ratio. The unusually large energy divisor makes that aggregate channel close to zero at typical energies and is an implementation limitation. The same six-value input is returned to every cell with that \texttt{species\_id}. Each cell then applies its own stored 6 $\times$ 4 linear--tanh strategy head and separate 6 $\times$ 2 linear--tanh movement head. Strategy is concatenated into W2 and movement feeds the velocity update. These weights are per cell and inherited from the reproducing parent. They mutate only when global evolution is enabled; there is no context-head RL update path, so EvoEvo and EvoRL can mutate them during training whereas RLRL cannot. They are not one shared network object for the species.

\begin{figure*}[t!]
\centering
\includegraphics[width=0.96\textwidth]{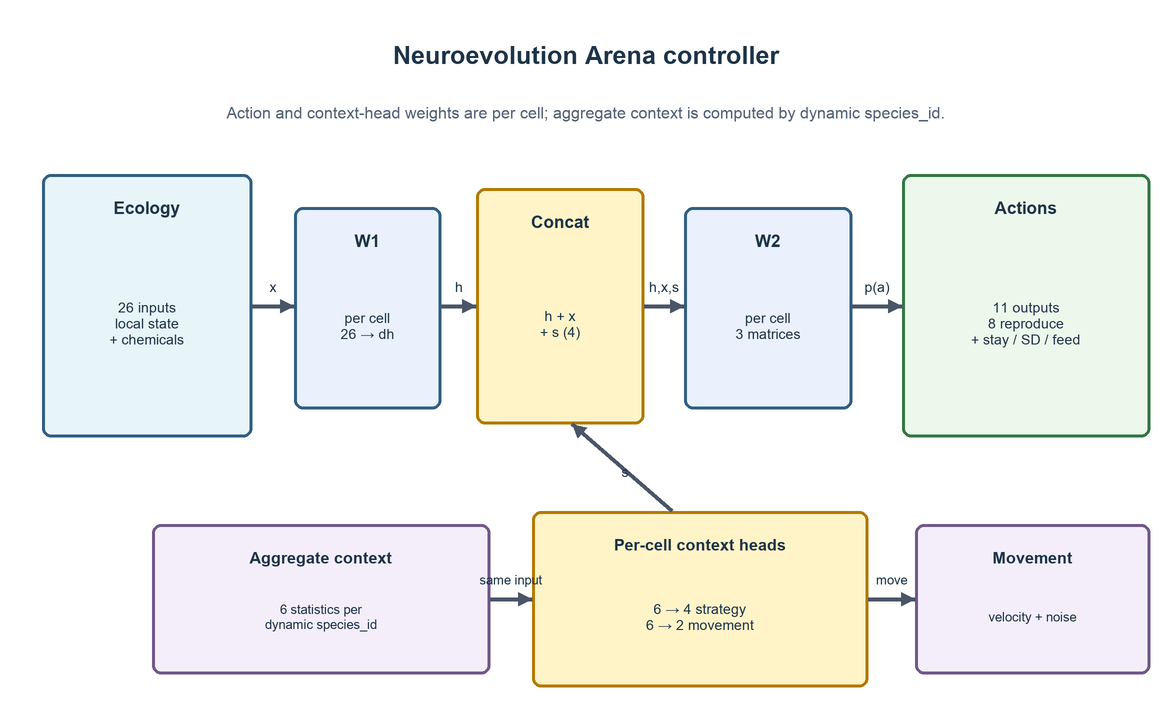}
\caption{Implementation-faithful system overview. Every cell stores its own W1/W2 action-network weights and its own context strategy and movement-head weights. Cells sharing an explicit \texttt{species\_id} receive the same six aggregate statistics, but their inherited context-head parameters need not be identical. The four-dimensional strategy enters W2; the two-dimensional movement output drives a separate Boid-style velocity update.}\label{fig:controller}
\end{figure*}

At each step, the movement update multiplies velocity by 0.7 and adds 0.3 times the sum of the cell-local movement-head output and Gaussian noise with standard deviation 0.1. Velocity is clipped to {[}-1, 1{]}, multiplied by a maximum step of 3, and rounded to a grid displacement; uncontested moves into empty toroidally wrapped targets cost 0.5 energy. The legacy artifacts store selected action-network weights but not the context strategy head, movement head, full population, or ecological state. Formal evaluation reconstructs these omitted fields from the seed-defined evaluation context under one standardized policy applied to every condition and freezes controller and lineage-template values. Population state and dynamic \texttt{species\_id} labels continue to change through ecological processes. The experiment therefore embeds frozen cell-network weights in a standardized new ecology rather than replaying a complete trained ecosystem. No context-head or movement ablation was performed. Figure~\ref{fig:controller} summarizes this implementation-faithful controller and context flow.

\subsection{Competition Protocol}\label{competition-protocol}

Frozen evaluation uses one run-level saved elite-controller artifact from each of the 18 runs in the replicated 3 $\times$ 2 training experiment: six update-regime--architecture conditions, each represented by three training runs. Each artifact contains selected best-W1/W2 action-network matrices recorded from one cell within its run; it is neither the terminal population nor a complete ecosystem snapshot. No training run was omitted through cross-run best-run selection, but within-run elite selection remains. Lifetime learning, evolutionary crossover or averaging, mutation, output-layer reset, alien injection, and tissue-fission mutation are disabled and audited. Births clone the canonical action network, context heads, chemical affinity, cooperation scalar, model type, architecture, and model origin from one parent exactly. Ecological state is not static: population changes, movement, resource dynamics, and the 50-step disconnected-component split of dynamic \texttt{species\_id} remain active. Every job starts from 1,200 cells on the 100 $\times$ 100 grid. Pairwise jobs allocate 600 cells to each condition, six-way jobs allocate 200 cells to each condition, and survival jobs use one 1,200-cell condition. Pairwise and six-way jobs run for 5,000 steps; population is recorded every 50 steps and AUC aggregated by immutable \texttt{model\_origin} is the primary competition metric.

The \emph{aligned-run frozen-evaluation design} contains all 15 unordered condition pairs $\times$ three matched training-run indices $\times$ three seed-defined ecological contexts = 135 pairwise jobs. Thus run 0 competes with run 0, and so forth; the design does not evaluate the full $3\times3$ cross-run artifact combinations for each condition pair. For condition A versus B, the primary effect is $\log((\mathrm{AUC}_A+1)/(\mathrm{AUC}_B+1))$. The three context effects are equally averaged within an aligned training-run block, after which the three blocks form the independent descriptive level.

The three nominal context seed codes are not exchangeable repetitions of one environment. Codes 20260727 and 20260728 initialize every condition with a positive cooperation scalar, whereas code 20260729 initializes every condition with a non-positive scalar and permits energy-based attack. The pooled block effect consequently weights cooperation-permitting contexts 2:1 relative to the attack-permitting context, in addition to changing randomized spatial and ancillary state. We retain this prespecified equal-context average but interpret it only as a summary of that particular context mixture. The nine six-way jobs use the same three-by-three artifact/context structure and population AUC, but remain an exploratory context probe rather than a second estimate of the pairwise effect.

The 54 survival jobs cover six conditions $\times$ three training-run artifacts $\times$ three seed-defined contexts. Each job applies a 97\% shock and allows 1,000 recovery steps; recovery requires population $\geq$ 5,000. The 97\% level may be attempted up to three times. A job proceeds to 98\%, then to repeated 99\% shocks, only after reaching the threshold at the preceding level. The primary survival endpoint is the maximum shock ratio from which threshold recovery occurs. Across the complete frozen evaluation there are 198 computational jobs, but the independent training-run level remains n = 3 per condition.

Table~\ref{tab:3} distinguishes simulation-job counts from the independent training-run level and summarizes the formal frozen-evaluation protocol. Unless otherwise noted, training constants were selected through pilot development; the aligned-run design and primary effects were fixed before the reported batch.

\begin{table*}[t!]
\centering
\small
\caption{Key experimental and frozen-evaluation parameters. The 198 jobs are computational evaluations, not 198 independent statistical replicates.}\label{tab:3}
\resizebox{\textwidth}{!}{%
\begin{tabular}{@{}
  >{\raggedright\arraybackslash}p{(\linewidth - 2\tabcolsep) * \real{0.4846}}
  >{\raggedright\arraybackslash}p{(\linewidth - 2\tabcolsep) * \real{0.5154}}@{}}
\toprule\noalign{}
\begin{minipage}[b]{\linewidth}\raggedright
\textbf{Parameter}
\end{minipage} & \begin{minipage}[b]{\linewidth}\centering
\textbf{Value}
\end{minipage} \\
\midrule\noalign{}
\multicolumn{2}{@{}>{\raggedright\arraybackslash}p{(\linewidth - 2\tabcolsep) * \real{1.0000} + 2\tabcolsep}@{}}{%
\textbf{Environment}} \\
Grid size & 100 $\times$ 100 (toroidal) \\
Metabolic scaling onset & 800 cells \\
Hunger death threshold & 150 \\
Baseline metabolic rate & 0.05 energy/step \\
Density scaling parameter & 130 cells \\
Isolation hunger penalty & +10/step when \texttt{n\_kin} \textless{} 2 \\
Overcrowding hunger penalty & +10/step per kin above 6 \\
Reproduction energy threshold & 8 \\
Chemical kernel / blend / decay & fixed 7$\times$7 / 0.3 / 2\% per step (zero padded) \\
\multicolumn{2}{@{}>{\raggedright\arraybackslash}p{(\linewidth - 2\tabcolsep) * \real{1.0000} + 2\tabcolsep}@{}}{%
\textbf{Training}} \\
Generations & 50,000 (10,000 in horizon probe) \\
Runs per condition & 3 (seeds 42, 43, 44) \\
RL learning rate $\eta$ & 0.01 \\
Experience replay buffer & 10K \\
\multicolumn{2}{@{}>{\raggedright\arraybackslash}p{(\linewidth - 2\tabcolsep) * \real{1.0000} + 2\tabcolsep}@{}}{%
\textbf{Competition}} \\
Frozen match length & 5,000 steps \\
Initial total population & 1,200 (600 per pairwise condition; 200 per six-way condition) \\
Aligned-run pairwise design & 15 pairs $\times$ 3 matched run indices $\times$ 3 contexts = 135 jobs \\
Six-way / survival matrix & 9 / 54 jobs (198 frozen-evaluation jobs total) \\
Survival recovery rule & 1,000 steps; threshold 5,000; 97\% first (up to 3 attempts) \\
Seed-defined contexts & 20260727 and 20260728: cooperation-permitting; 20260729: attack-permitting \\
Context aggregation & equal mean, hence cooperation:attack weighting of 2:1 \\
Independent analysis level & 3 aligned training-run blocks; contexts nested within block \\
Auxiliary horizon probe & unmatched 10K single run and 50K three-run batch; reported in the appendix \\
\bottomrule
\end{tabular}
}
\end{table*}

\section{Replicated Update-Regime--Architecture Evaluation}\label{replicated-update-regimearchitecture-evaluation}

\subsection{Setup}\label{setup}

The main experiment combines replicated training with the formal 3 $\times$ 2 aligned-run frozen-evaluation design. EvoEvo, EvoRL, and RLRL are each trained with MLP4 and Wide128 for 50,000 generations in three independent runs (nominal seeds 42, 43, and 44), yielding 18 run-level saved elite-controller artifacts. Table~\ref{tab:4} reports the MLP4 subset to isolate update-regime differences in training; factorial contrasts are discussed later. One artifact from every run enters the frozen protocol described above. The resulting 198 jobs comprise 135 pairwise competitions, 9 six-way melees, and 54 survival jobs. The pairwise analysis first averages the three seed-defined context effects within each aligned run block and treats the three blocks as the independent descriptive level.

\subsection{Training Results}\label{training-results}

\begin{table*}[t!]
\centering
\small
\caption{Training performance by regime for MLP4 at 50K generations (three runs).}\label{tab:4}
\resizebox{\textwidth}{!}{%
\begin{tabular}{@{}
  >{\raggedright\arraybackslash}p{(\linewidth - 6\tabcolsep) * \real{0.1923}}
  >{\centering\arraybackslash}p{(\linewidth - 6\tabcolsep) * \real{0.3462}}
  >{\centering\arraybackslash}p{(\linewidth - 6\tabcolsep) * \real{0.2308}}
  >{\centering\arraybackslash}p{(\linewidth - 6\tabcolsep) * \real{0.2308}}@{}}
\toprule\noalign{}
\begin{minipage}[b]{\linewidth}\raggedright
\textbf{Regime}
\end{minipage} & \begin{minipage}[b]{\linewidth}\centering
\textbf{Fitness (mean $\pm$ std)}
\end{minipage} & \begin{minipage}[b]{\linewidth}\centering
\textbf{Population}
\end{minipage} & \begin{minipage}[b]{\linewidth}\centering
\textbf{Lifetime}
\end{minipage} \\
\midrule\noalign{}
EvoEvo & 149,957 $\pm$ 7,600 & 6,008 & 29.1 \\
EvoRL & 335,371 $\pm$ 54,635 & 5,071 & 67.1 \\
RLRL & 299,081 $\pm$ 48,493 & 4,905 & 62.8 \\
\bottomrule
\end{tabular}
}
\end{table*}

Both RL-enabled regimes had higher recorded training fitness than EvoEvo in the MLP4 subset: EvoRL reached 335K and RLRL 299K, compared with 150K for EvoEvo. The simulator's cell-level fitness is lifetime + 10 $\times$ \texttt{reproduction\_count} + 0.5 $\times$ energy/100 for living cells. Diversity is disabled in the metric calls used here, and population size is reported separately rather than entering this formula directly. Training fitness is therefore an implementation-specific individual score, not a direct population-level robustness measure.

The component statistics show contrasting population profiles. EvoEvo maintained the largest population (6,008) but the shortest lifetime (29.1 steps), whereas EvoRL and RLRL maintained smaller populations (5,071 and 4,905) with longer lifetimes (67.1 and 62.8). Run-to-run variability was also higher for the two RL-enabled conditions (both about 16\% of the mean) than for EvoEvo (about 5\%). These are descriptive differences; possible mechanisms and population--lifetime tradeoffs are considered in Section 6.

\subsection{Formal Frozen Competition Results}\label{formal-frozen-competition-results}

For each of the 15 unordered condition pairs, we evaluated artifacts with corresponding run indices under three seed-defined ecological contexts, giving 15 $\times$ 3 $\times$ 3 = 135 pairwise jobs. The primary effect for condition A versus B is $\log((\mathrm{AUC}_A+1)/(\mathrm{AUC}_B+1))$. Context effects are averaged within each aligned run block before comparing directions across the three blocks. Because two contexts permit cooperation and one permits attack, this pooled effect is an explicitly protocol-weighted summary rather than a neutral average over exchangeable random seeds. Raw job-level wins are descriptive and are not treated as n = 135 independent observations.

EvoEvo$\times$MLP4 had the largest pairwise run-block count under this protocol. It won 11 of its 15 opponent-by-aligned-run blocks and held a 2-of-3 or 3-of-3 run-block majority against each of its five opponents. EvoRL$\times$MLP4, RLRL$\times$MLP4, and RLRL$\times$Wide128 each won 9 of 15 blocks; EvoRL$\times$Wide128 won 5, and EvoEvo$\times$Wide128 won 2. Table~\ref{tab:5} summarizes these counts across the six conditions. They characterize the specified 2:1 mixture of cooperation- and attack-permitting contexts and are not a context-independent strength scale.

\begin{table*}[t!]
\centering
\small
\caption{Aligned-run pairwise summary by condition. A block win is the positive AUC log-ratio direction after equally averaging two cooperation-permitting contexts and one attack-permitting context within one matched run index. Opponent majorities count how many of the other five conditions were beaten in at least two of three aligned blocks.}\label{tab:5}
\resizebox{\textwidth}{!}{%
\begin{tabular}{@{}
  >{\raggedright\arraybackslash}p{(\linewidth - 4\tabcolsep) * \real{0.3308}}
  >{\centering\arraybackslash}p{(\linewidth - 4\tabcolsep) * \real{0.2692}}
  >{\centering\arraybackslash}p{(\linewidth - 4\tabcolsep) * \real{0.4000}}@{}}
\toprule\noalign{}
\begin{minipage}[b]{\linewidth}\raggedright
\textbf{Condition}
\end{minipage} & \begin{minipage}[b]{\linewidth}\centering
\textbf{Run-block wins (of 15)}
\end{minipage} & \begin{minipage}[b]{\linewidth}\centering
\textbf{Opponent majorities (of 5)}
\end{minipage} \\
\midrule\noalign{}
EvoEvo$\times$MLP4 & 11 & 5 \\
EvoEvo$\times$Wide128 & 2 & 0 \\
EvoRL$\times$MLP4 & 9 & 3 \\
EvoRL$\times$Wide128 & 5 & 1 \\
RLRL$\times$MLP4 & 9 & 3 \\
RLRL$\times$Wide128 & 9 & 3 \\
\bottomrule
\end{tabular}
}
\end{table*}

With n = 3 aligned training-run blocks, the descriptive majority ordering reverses across architectures. Within MLP4, EvoEvo beat EvoRL by 2 blocks to 1, EvoEvo beat RLRL 2--1, and EvoRL beat RLRL 2--1, giving EvoEvo \textgreater{} EvoRL \textgreater{} RLRL. Within Wide128, RLRL beat EvoRL 2--1 and EvoEvo 3--0, while EvoRL beat EvoEvo 2--1, giving RLRL \textgreater{} EvoRL \textgreater{} EvoEvo. In addition, MLP4 beat Wide128 by 2--1 within each regime. These are sampled majority directions under the aligned, context-weighted protocol, not estimated population rankings.

These majority orderings should not be read as highly stable rankings. Only 4 of the 15 condition pairs retained the same effect direction across all three aligned run blocks; the other 11 reversed direction in one block. At the nested-context level, only 17 of 45 pair-by-run blocks produced the same winner under all three contexts. Because the context identifier changes both randomized ecological state and the categorical contact rule, these disagreements cannot be interpreted as ordinary seed variance. The complete pooled directions are reported in Table~\ref{tab:6}. The evidence supports architecture-conditioned outcomes under the prespecified three-context mixture and substantial artifact dependence, not a deterministic global hierarchy. Context-specific effects should be estimated in a future balanced design rather than reverse-engineered from this 2:1 mixture.

Aggregating only cross-regime contests gives RLRL the largest positive block count (15 positive versus 9 negative), compared with EvoRL (11--13) and EvoEvo (10--14). This aggregation does not establish a universal regime rank: EvoEvo$\times$MLP4 has 11 condition-level wins while EvoEvo$\times$Wide128 has 2, so collapsing over architecture hides the largest contrast. Raw job wins and condition-pair majorities also produce different secondary orderings and do not replace the prespecified run-block effect.

\begin{figure*}[t!]
\centering
\includegraphics[width=0.96\textwidth]{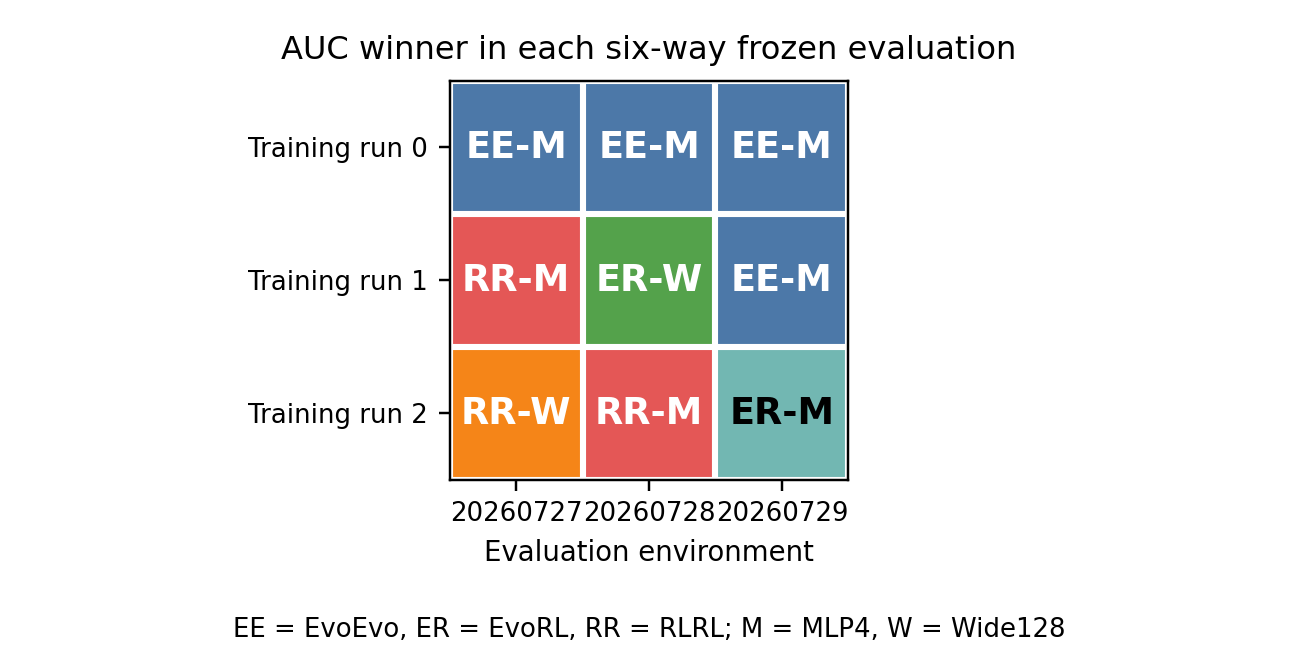}
\caption{Winners of the nine six-way frozen evaluations across three aligned artifact blocks and three nested seed-defined contexts. Five different conditions win at least once; only block 0 has the same winner across all three contexts. Winner is defined by population AUC. EE = EvoEvo, ER = EvoRL, RR = RLRL; M = MLP4, W = Wide128.}\label{fig:winners}
\end{figure*}

The nine six-way jobs were even more diverse in their winners, but they were retained as an exploratory context probe rather than used to claim a separate stability coefficient. EvoEvo$\times$MLP4 won 4 of 9, RLRL$\times$MLP4 won 2, and EvoRL$\times$MLP4, EvoRL$\times$Wide128, and RLRL$\times$Wide128 each won 1; EvoEvo$\times$Wide128 won none. Aligned run block 0 produced the same winner under all three contexts, whereas blocks 1 and 2 each produced three different winners. Eight of nine final states contained only one surviving origin. Because artifact identity, randomized ecological state, and contact-rule context all vary, this diversity cannot be assigned to any one source (Figure~\ref{fig:winners}).

\begin{table*}[t!]
\centering
\small
\caption{Aligned-run pairwise direction matrix (wins out of three matched run-index blocks). Each cell is based on the mean AUC log-ratio over the three nested contexts, with a 2:1 cooperation-to-attack weighting. EE = EvoEvo, ER = EvoRL, RR = RLRL; M4 = MLP4 and W = Wide128.}\label{tab:6}
\resizebox{\textwidth}{!}{%
\begin{tabular}{@{}
  >{\raggedright\arraybackslash}p{(\linewidth - 12\tabcolsep) * \real{0.2285}}
  >{\centering\arraybackslash}p{(\linewidth - 12\tabcolsep) * \real{0.1286}}
  >{\centering\arraybackslash}p{(\linewidth - 12\tabcolsep) * \real{0.1286}}
  >{\centering\arraybackslash}p{(\linewidth - 12\tabcolsep) * \real{0.1286}}
  >{\centering\arraybackslash}p{(\linewidth - 12\tabcolsep) * \real{0.1286}}
  >{\centering\arraybackslash}p{(\linewidth - 12\tabcolsep) * \real{0.1286}}
  >{\centering\arraybackslash}p{(\linewidth - 12\tabcolsep) * \real{0.1286}}@{}}
\toprule\noalign{}
\begin{minipage}[b]{\linewidth}\raggedright
\textbf{Condition}
\end{minipage} & \begin{minipage}[b]{\linewidth}\centering
\textbf{EE$\times$M4}
\end{minipage} & \begin{minipage}[b]{\linewidth}\centering
\textbf{EE$\times$W}
\end{minipage} & \begin{minipage}[b]{\linewidth}\centering
\textbf{ER$\times$M4}
\end{minipage} & \begin{minipage}[b]{\linewidth}\centering
\textbf{ER$\times$W}
\end{minipage} & \begin{minipage}[b]{\linewidth}\centering
\textbf{RR$\times$M4}
\end{minipage} & \begin{minipage}[b]{\linewidth}\centering
\textbf{RR$\times$W}
\end{minipage} \\
\midrule\noalign{}
EE$\times$M4 & --- & 2 & 2 & 3 & 2 & 2 \\
EE$\times$W & 1 & --- & 0 & 1 & 0 & 0 \\
ER$\times$M4 & 1 & 3 & --- & 2 & 2 & 1 \\
ER$\times$W & 0 & 2 & 1 & --- & 1 & 1 \\
RR$\times$M4 & 1 & 3 & 1 & 2 & --- & 2 \\
RR$\times$W & 1 & 3 & 2 & 2 & 1 & --- \\
\bottomrule
\end{tabular}
}
\end{table*}

In short, Tables~\ref{tab:5} and~\ref{tab:6} show the central empirical result: the apparent ordering of update regimes depends on architecture, and the limited directional agreement across aligned artifacts prevents any condition from being treated as a universal competitive leader.

\subsection{Replicated Frozen Survival Results}\label{replicated-frozen-survival-results}

The formal survival design contains 54 jobs: six conditions $\times$ three training-run artifacts $\times$ three seed-defined contexts. Each job begins with one 1,200-cell condition, applies a 97\% population shock, and allows 1,000 recovery steps. Recovery is defined a priori as reaching a population of 5,000. A job may attempt the 97\% level up to three times; it advances to 98\% and then 99\% only after threshold recovery at the preceding level. The primary endpoint is the maximum shock ratio from which threshold recovery is achieved.

The primary endpoint produced a complete floor effect. None of the 54 jobs reached the 5,000-cell threshold after any 97\% attempt, so every condition had a maximum recovered ratio of 0. No job advanced to the planned 98\% or 99\% stages. Fifty jobs remained non-extinct after their attempted 97\% shocks, while four became extinct; non-extinction below threshold is not counted as recovery under the prespecified endpoint.

A secondary, descriptive endpoint---the final population after attempted shocks---was highest for EvoRL$\times$MLP4 (mean of training-run means 2,430 $\pm$ 410), followed by RLRL$\times$Wide128 (2,194 $\pm$ 383), EvoEvo$\times$MLP4 (2,121 $\pm$ 226), RLRL$\times$MLP4 (2,036 $\pm$ 182), EvoEvo$\times$Wide128 (1,697 $\pm$ 332), and EvoRL$\times$Wide128 (1,619 $\pm$ 736). EvoRL$\times$MLP4 led the aligned comparison in only two of three training-run blocks. Because this endpoint was secondary and the primary score is tied at zero, it is reported as exploratory rather than used to rank survival robustness.

\subsection{Cross-Metric Synthesis}\label{cross-metric-synthesis}

The metrics answer different questions and do not support one composite leaderboard. Recorded training fitness favors the RL-enabled configurations, the largest aligned-run pairwise count belongs to EvoEvo$\times$MLP4, five conditions win at least one six-way job, and the survival primary endpoint cannot distinguish any condition. The architecture-conditioned pairwise directions are the main ecological pattern; their limited agreement across aligned artifacts is part of the result rather than a nuisance to be averaged away.

\section{Discussion}\label{discussion}

\subsection{Layer-Partitioned Evolution and RL: Limits of the Baldwin Analogy}\label{layer-partitioned-evolution-and-rl-limits-of-the-baldwin-analogy}

EvoRL was motivated by the Baldwin Effect, but the implemented cross-experiment is not a classical Baldwin or sequential evolution-then-learning protocol. It has no W2 weight update for the first 3\% of training, although reward can accumulate during that interval; thereafter W1 evolution and W2 lifetime updates operate concurrently. The split is prescribed, not evolved, and RL-modified W2a/W2b values are directly inherited or averaged by offspring, which is Lamarckian-like. Formal evaluation then freezes all controller updates. Competition and recovery therefore measure ecological consequences of trained weights, not learning after an opponent encounter or disaster, and cannot demonstrate a Baldwin mechanism.

The formal evidence also does not identify EvoRL as uniquely competitive or resilient across architectures. A biological analogy remains useful only as a hypothesis that layer-specific update assignments may create different inductive biases. Demonstrating Baldwinian dynamics would require explicit ablations of the 3\% warmup and layer assignment, an evolved or varied learning schedule, and evaluation that measures learning-dependent adaptation. We therefore describe EvoRL as a \emph{Baldwin-inspired layer-partitioned regime}, never as evidence of a Baldwin Effect.

\subsection{Replicated Update-Regime--Architecture Interaction}\label{replicated-update-regimearchitecture-interaction}

As the core replicated analysis of RQ1 and RQ2, we crossed all three regimes with MLP4 and Wide128 and trained every condition for 50,000 generations in three runs. The formal follow-up evaluated one saved artifact from all 18 runs: no run was omitted through cross-run best-run selection, although each artifact contains action-network weights selected from an elite cell within that run. It contains 135 pairwise jobs, 9 six-way jobs, and 54 survival jobs. Seed-defined contexts are nested within three aligned run-index blocks, so 198 is the number of computational jobs rather than the statistical sample size. The design covers all available artifacts but not all cross-run pairings.

In training fitness, the update-regime main-effect contrast was about 11 times larger than the tested architecture main-effect contrast: moving from EvoEvo to either RL-enabled regime changed fitness by roughly 106,000 relative to the grand mean, whereas the MLP4--Wide128 contrast was roughly 9,400. This ratio is specific to the current 3 $\times$ 2 design, fitness definition, architectures, and training batch; it is not an estimate of a universal update-to-architecture effect ratio.

The training analysis also showed a crossover interaction. EvoRL had a positive interaction term with MLP4 (+21,504 relative to the additive prediction) and a negative term with Wide128 (-21,504), while RLRL showed the opposite pattern (+16,209 for Wide128 and -16,209 for MLP4). These coefficients describe training fitness. They motivate, but do not predetermine, the ranking under ecological competition. Table~\ref{tab:factorial} summarizes these descriptive factorial patterns.

\begin{table*}[t!]
\centering
\small
\caption{Descriptive factorial summary for the 3 $\times$ 2 update-regime--architecture training analysis (three runs per condition).}\label{tab:factorial}
\resizebox{\textwidth}{!}{%
\begin{tabular}{@{}
  >{\raggedright\arraybackslash}p{(\linewidth - 4\tabcolsep) * \real{0.3538}}
  >{\centering\arraybackslash}p{(\linewidth - 4\tabcolsep) * \real{0.2692}}
  >{\centering\arraybackslash}p{(\linewidth - 4\tabcolsep) * \real{0.3769}}@{}}
\toprule\noalign{}
\begin{minipage}[b]{\linewidth}\raggedright
\textbf{Pattern}
\end{minipage} & \begin{minipage}[b]{\linewidth}\centering
\textbf{Value}
\end{minipage} & \begin{minipage}[b]{\linewidth}\centering
\textbf{Interpretation}
\end{minipage} \\
\midrule\noalign{}
Update-regime effect & \textasciitilde106K fitness & larger descriptive contrast \\
Architecture effect & \textasciitilde9.4K fitness & smaller descriptive contrast \\
EvoRL$\times$MLP4 & +21,504 fitness & positive deviation from additive mean \\
EvoRL$\times$Wide128 & -21,504 fitness & negative deviation from additive mean \\
RLRL$\times$Wide128 & +16,209 fitness & positive deviation from additive mean \\
RLRL$\times$MLP4 & -16,209 fitness & negative deviation from additive mean \\
\bottomrule
\end{tabular}
}
\end{table*}

One hypothesis for the training interaction is that coordination between evolution-updated W1 representations and RL-updated W2 policies is easier in the smaller MLP4 parameter space, whereas RL on both layer groups can exploit the greater capacity of Wide128. The formal competition results do not reproduce this explanation as a simple pairwise hierarchy: MLP4 beats Wide128 within all three regimes by 2--1 aligned-block majorities, and EvoEvo$\times$MLP4 has the largest pairwise block count. The proposed search-space mechanism is therefore speculative and would require layer-specific representation and optimization diagnostics.

The pairwise summaries show architecture-conditioned regime patterns in the sampled artifacts. The majority order is EvoEvo \textgreater{} EvoRL \textgreater{} RLRL within MLP4 but RLRL \textgreater{} EvoRL \textgreater{} EvoEvo within Wide128. The largest descriptive contrast is inside EvoEvo: MLP4 has condition-pair majorities against all five opponents, whereas Wide128 has none. Because only four condition pairs are directionally unanimous across aligned blocks, the result is reported as a protocol-specific pattern with substantial artifact-to-artifact uncertainty rather than a universal rank order.

Across evaluation dimensions, no single condition dominates. Training fitness is highest for RLRL$\times$Wide128, EvoEvo$\times$MLP4 has the largest aligned-run pairwise count, and the nine six-way jobs distribute wins across five conditions. The strict survival endpoint yields no winner because all 54 scores are zero. EvoRL$\times$MLP4 has the largest secondary final-population summary after shocks. This pattern shows that training and evaluation protocols emphasize different properties and that an exploratory secondary endpoint must not be promoted into a global ranking.

\subsection{Evaluation Reliability and Context Dependence}\label{evaluation-reliability-and-context-dependence}

Evaluation context and replication level shape the apparent ranking. Training fitness, pairwise AUC, six-way exclusion, and shock recovery measure different properties; they should not be collapsed into one leaderboard. Artifact variation is also distinct from context variation. The present design partially exposes both, but the context factor is confounded: two seed codes instantiate cooperation-permitting contact and one instantiates attack-permitting contact. Equal averaging therefore represents a chosen 2:1 ecological mixture, not replication from one exchangeable seed distribution.

Pairwise decomposition remains useful for opponent-specific description, but the aligned-run design leaves an important gap. It compares only artifacts with the same run index, so an apparent condition effect can depend on three arbitrary pairings. A full cross-run design would evaluate all nine artifact combinations per condition pair and separate focal-artifact, opponent-artifact, and ecological-context contributions. The current 4-of-15 unanimous pair directions and 17-of-45 context-consistent pair-by-run blocks justify uncertainty, but do not by themselves identify which source causes a reversal.

We therefore propose four evaluation practices: (1) replicate training runs as the independent level; (2) cross saved artifacts where feasible instead of relying only on aligned indices; (3) represent qualitatively different ecological contexts as an explicit factor and balance them before pooling; and (4) specify primary endpoints before a batch and report null or floor effects rather than selecting a favorable secondary metric after inspection. Pairwise, multi-way, and survival protocols remain complementary because they test different ecological properties; each requires its own replicated, resolving endpoint.

\subsection{Survival Floor and Interpretive Boundaries}\label{survival-floor-and-interpretive-boundaries}

The strict survival endpoint is non-discriminating: all six conditions fail to recover to 5,000 cells after the 97\% shock in every artifact--context combination. This negative result is informative about protocol severity, but it contains no evidence for a relative survival ranking and no observation at the planned 98\% or 99\% stages.

EvoRL$\times$MLP4 has the highest secondary mean final population, but that measure is not the primary recovery endpoint and its direction varies across training runs. It cannot support a specialist--generalist classification, an r/K-like interpretation, or a Baldwin-like resilience claim. Re-thresholding the present 54 jobs after inspection would convert an exploratory rescue analysis into an unplanned primary claim.

Future survival work should calibrate thresholds on separate pilot data and then preregister multiple interpretable endpoints, such as time to a fixed fraction of pre-shock population, recovery AUC, extinction probability, and survivor diversity. Those endpoints require new evaluation data; the current floor-affected batch should remain reported as a null primary result.

\subsection{Limitations}\label{limitations}

Several limitations constrain generalization. The primary factorial covers three regimes and two feed-forward architectures; Hybrid64-128 and NoSkip64 occur only in the exploratory appendix. Each condition has three independent training runs. The 198 evaluation jobs improve coverage, but nested contexts do not create n = 198 independent replicates. Pairwise artifacts are matched only by run index rather than evaluated in a full cross-run design.

The legacy artifacts contain within-run elite-selected action-network weights rather than complete saved ecosystems. Training also combines imperfectly shared randomness with regime-specific inheritance and ancillary-state behavior; known mate-search and warmup-reward defects remain in the data-generating implementation. The results therefore characterize these saved artifacts and do not isolate a clean causal effect of lifetime RL.

Formal evaluation reconstructs omitted ecological state under three seed-defined contexts. Two contexts permit cooperation and one permits attack, so context and nominal seed are confounded and the equal-weight mean favors cooperation contexts 2:1. Dynamic species identities, contact rules, and population fragmentation remain protocol-dependent, and the prespecified survival threshold produces a complete floor before the planned 98\% and 99\% stages. Accordingly, competitive and survival conclusions remain conditional on the aligned pairings, context mixture, implemented identity rules, and threshold choice.

\subsection{Future Work}\label{future-work}

Immediate work should add independent training runs and evaluate all cross-run artifact pairings rather than only aligned indices. Ecological contact mode should be manipulated independently from random initialization, with balanced cooperation- and attack-permitting contexts reported both separately and through a prespecified aggregate. The four-architecture probe requires corrected frozen competition before any ecological ranking of Hybrid64-128 or NoSkip64 is reported. A new preregistered survival study should calibrate its threshold and recovery horizon on separate pilot data, then use recovery-curve endpoints capable of distinguishing persistent subthreshold populations without redefining success post hoc. Future saved artifacts should include the complete population, model type, species state, configuration, and RNG provenance. Sensitivity studies should vary grid scale, metabolism, reproduction, chemical/combat dynamics, and dominance control. Longer-term work can co-evolve architectures and layer-update schedules and directly measure adaptation during evaluation.

\section{Conclusion}\label{conclusion}

Neuroevolution Arena provides a shared ecology in which update-and-inheritance regimes and neural architectures can be compared during training and under frozen ecological evaluation. In the tested 3 $\times$ 2 design, higher recorded training fitness does not determine the largest pairwise count, and the pairwise majority direction of the three regimes changes with architecture. Six-way outcomes vary across artifacts and contexts, while the prespecified survival threshold yields a complete floor. These results support a bounded claim: ecological outcomes in this system depend jointly on controller configuration, saved training-run artifact, opponent set, and evaluation context; they do not identify a universally superior regime, a survival leader, or a demonstrated Baldwin Effect.

The study contributes an audit-tracked nested protocol that makes those dependencies explicit and cleanly separates training-run, architecture, and evaluation-context levels of variation. Its boundary is equally important: 198 jobs do not provide 198 independent replicates, the present pairwise design aligns rather than fully crosses training-run artifacts, and the three context codes encode a 2:1 mixture of cooperation- and attack-permitting contexts. Future evaluations should cross artifacts, manipulate contact context independently from random state, and use resolving prespecified endpoints. Treating these design levels separately is more informative than forcing training, pairwise, multi-way, and recovery outcomes into one score.

\paragraph{Artifact availability.} Source code is publicly available at \url{https://github.com/geyuxu/neuroevolution-arena}. Trained models are hosted at \url{https://huggingface.co/geyuxu/alife2026-model}, and evaluation datasets at \url{https://huggingface.co/datasets/geyuxu/alife2026-data}. A versioned source-code release should be archived with a persistent identifier before journal publication.

\paragraph{Declaration of generative-AI assistance.} During manuscript preparation, the authors used OpenAI Codex for language editing, restructuring, LaTeX conversion, code-assisted consistency checks, and drafting revisions. The authors reviewed and verified the resulting text, analyses, citations, and claims and take full responsibility for the manuscript.

\appendix
\section{Unmatched Architecture Training-Horizon Probe}\label{appendix:horizon-probe}

This auxiliary dataset is retained for provenance but is not part of the main research questions. It compares Baseline64 (MLP4; $d_h=d_r=64$), Wide128 ($d_h=d_r=128$), Hybrid64-128 ($d_h=64$, $d_r=128$), and NoSkip64 ($d_h=d_r=64$ without the skip connection), all under EvoRL. Each architecture was trained once to 10K generations and three times to 50K in a separate batch. Table~\ref{tab:aux-horizon} reports training fitness and ranks for both batches. Consequently, horizon is confounded with batch and replication count; the values cannot establish improvement, decline, convergence rate, or a causal horizon effect. The legacy downstream competition and survival outputs are excluded because the audited runner did not maintain frozen weights and heritable state or restore model type correctly.

\begin{table*}[t!]
\centering
\small
\caption{Training fitness in the unmatched auxiliary architecture batches. The 10K column is a single run; the 50K column is mean $\pm$ standard deviation over three runs. These values must not be pooled with the main 3 $\times$ 2 experiment.}\label{tab:aux-horizon}
\begin{tabular}{@{}lrrrr@{}}
\toprule
Architecture & Fitness at 10K & 10K rank & Fitness at 50K & 50K rank \\
\midrule
Baseline64 (MLP4) & 336,851 & 1 & 282,362 $\pm$ 11,905 & 3 \\
Wide128 & 268,064 & 3 & 327,956 $\pm$ 38,210 & 1 \\
Hybrid64-128 & 309,143 & 2 & 308,342 $\pm$ 57,006 & 2 \\
NoSkip64 & 105,331 & 4 & 123,316 $\pm$ 24,195 & 4 \\
\bottomrule
\end{tabular}
\end{table*}

The rank difference between the single 10K batch and the three-run 50K batch is hypothesis-generating only. A valid horizon study would use matched independent replications at every horizon and would run the corrected frozen protocol for all four architectures before making ecological comparisons.

\clearpage
\footnotesize
\bibliographystyle{apalike}
\bibliography{references}

\end{document}